\documentclass[runningheads]{llncs}
\usepackage{graphicx}
\usepackage{url}            
\usepackage{booktabs}       
\usepackage{amsmath}
\usepackage{amssymb}
\usepackage{amsfonts}       
\usepackage{nicefrac}       
\usepackage{microtype}      
\usepackage{tikz}
\usepackage{tabularx}
\usepackage{hhline}
\usepackage{adjustbox}
\usepackage{multirow}
\usepackage{mathtools}
\usepackage{bbm}
\usepackage{bm}
\usepackage{color, colortbl}
\usepackage{pifont}
\usepackage{appendix}
\usepackage{diagbox}
\usepackage[misc]{ifsym}

\usepackage[colorlinks,linkcolor=red, anchorcolor=blue, citecolor=teal, urlcolor=magenta]{hyperref}

\usepackage{cellspace}
\DeclarePairedDelimiterX\set[1]\{\}{\nonscript\,#1\nonscript\,}

\usetikzlibrary{shapes.geometric, arrows}
\tikzstyle{startstop} = [rectangle, rounded corners, minimum width=1.5cm, minimum height=0.8cm,text centered, draw=black, fill=red!30]
\tikzstyle{io} = [trapezium, trapezium left angle=70, trapezium right angle=110, minimum width=1.5cm, minimum height=0.8cm, text centered, draw=black, fill=blue!30]
\tikzstyle{process} = [rectangle, minimum width=1.5cm, minimum height=0.8cm, text centered, draw=black, fill=orange!30]
\tikzstyle{decision} = [diamond, minimum width=1.5cm, minimum height=0.8cm, text centered, draw=black, fill=green!30]
\tikzstyle{block} = [rectangle, rounded corners, minimum width=1.5cm, minimum height=0.8cm,text centered, draw=black, fill=orange!30]
\tikzstyle{network} = [rectangle, rounded corners, minimum width=1.5cm, minimum height=0.8cm,text centered, draw=black, fill=green!30]
\tikzstyle{bim} = [rectangle, rounded corners, minimum width=0.1cm, minimum height=0.1cm, text centered, draw=blue, dash pattern=on 4pt off 4pt]
\tikzstyle{arrow} = [thick,->,>=stealth]

\DeclareMathOperator*{\argmin}{arg\,min}

\usepackage{makecell}

\newcommand\tf[1]{\textbf{#1}}

\def\ie{\textit{i.e.}}

\newcommand{\myparagraph}[1]{\vspace{1pt}\noindent{\bf{#1}}~~}
\makeatletter
\renewcommand{\paragraph}{%
  \@startsection{paragraph}{4}%
  {\z@}{0em}{-1em}%
  {\normalfont\normalsize\bfseries}%
}
\makeatother

\usepackage{colortbl}
\definecolor{lightgray}{gray}{0.75}
\definecolor{lightergray}{gray}{0.85}
\definecolor{Blue}{RGB}{3, 31, 97}
\definecolor{Blue1}{RGB}{214, 235, 245}
\definecolor{Blue2}{RGB}{235, 245, 250}
\definecolor{Gray}{RGB}{247, 252, 255}

\definecolor{convcolor}{HTML}{412F8A}
\definecolor{resnetcolor}{HTML}{8DA0CB}
\definecolor{vitcolor}{HTML}{fc8e62}

\newcommand{\convcolor}[1]{\textcolor{convcolor}{#1}}

\definecolor{aliceblue}{rgb}{0.94, 0.97, 1.0}

\newcommand{\cb}{\convcolor{$\bullet$\,}}
\newcommand{\gr}{\rowcolor[gray]{.95}}

\newcommand{\h}{{\bf h}}
\newcommand{\y}{{\bf y}}
\newcommand{\x}{{\bf x}}

\newcommand{\f}{{\bf f}}
\newcommand{\rr}{{\bf r}}
\newcommand{\uu}{{\bf u}}

\newcommand{\vv}{{\bf v}}
\newcommand{\w}{{\bf w}}

\begin{document}
\title{ACTION++: Improving Semi-supervised \\ Medical Image Segmentation with  Adaptive Anatomical Contrast}
\titlerunning{Adaptive Anatomical Contrast for Medical Image Segmentation}

\author{Chenyu You \textsuperscript{1(\Letter)} \and Weicheng Dai \inst{2} \and Yifei Min \inst{4} \and Lawrence Staib \inst{1,2,3} \and \\ Jas Sekhon \inst{4,5} \and James S. Duncan \inst{1,2,3,4}}
\authorrunning{C. You et al.}
\institute{\textsuperscript{1}Department of Electrical Engineering, Yale University
\\
\email{chenyu.you@yale.edu}\\
\textsuperscript{2}Department of Radiology and Biomedical Imaging, Yale University\\
\textsuperscript{3}Department of Biomedical Engineering, Yale University \\
\textsuperscript{4}Department of Statistics and Data Science, Yale University\\
\textsuperscript{5}Department of Political Science, Yale University\\
}

\maketitle              
\begin{abstract}
Medical data often exhibits long-tail distributions with heavy class imbalance, which naturally leads to difficulty in classifying the minority classes (\ie, boundary regions or rare objects). Recent work has significantly improved semi-supervised medical image segmentation in long-tailed scenarios by equipping them with unsupervised contrastive criteria. However, it remains unclear how well they will perform in the labeled portion of data where class distribution is also highly imbalanced. In this work, we present \tf{ACTION++}, an improved contrastive learning framework with adaptive anatomical contrast for semi-supervised medical segmentation. 
Specifically, we propose an adaptive supervised contrastive loss, where we first compute the optimal locations of class centers uniformly distributed on the embedding space (\ie, off-line), and then perform online contrastive matching training by encouraging different class features to adaptively match these distinct and uniformly distributed class centers. 
Moreover, we argue that blindly adopting a \textit{constant} temperature $\tau$ in the contrastive loss on long-tailed medical data is not optimal, and propose to use a \textit{dynamic} $\tau$ via a simple cosine schedule to yield better separation between majority and minority classes. Empirically, we evaluate ACTION++ on ACDC and LA benchmarks and show that it achieves state-of-the-art across two semi-supervised settings. Theoretically, we analyze the performance of adaptive anatomical contrast and confirm its superiority in label efficiency.

\end{abstract}

\section{Introduction}
\label{section:intro}

\begin{figure}[t]
\centering
\includegraphics[width=0.98\linewidth]{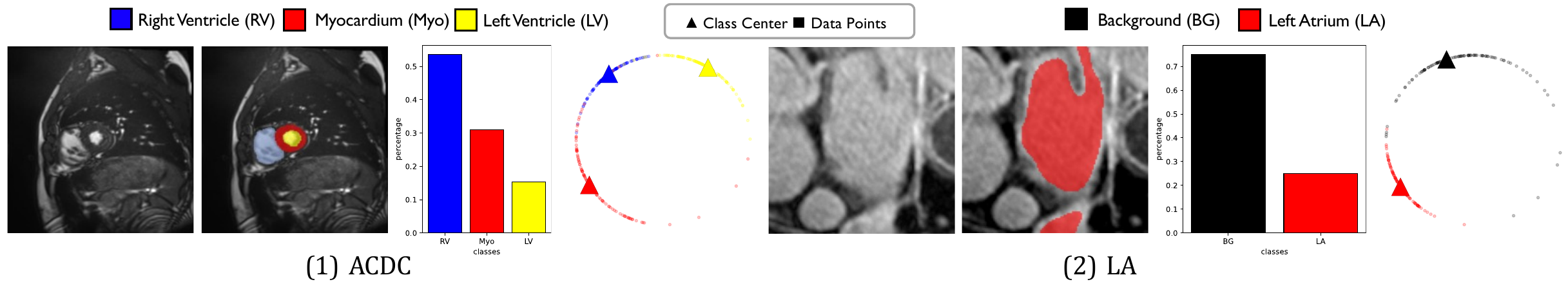}
\vspace{-12pt}
\caption{Examples of two benchmarks (\ie, ACDC and LA) with imbalanced class distribution. From left to right: input image, ground-truth segmentation map, class distribution chart, training data feature distribution for multiple classes.} 
\label{fig:distribution}
\vspace{-15pt}
\end{figure}

With the recent development of semi-supervised learning (SSL) \cite{chapelle2009semi}, rapid progress has been made in medical image segmentation, which typically learns rich anatomical representations from few labeled data and the vast amount of unlabeled data.  Existing SSL approaches can be generally categorized into adversarial training \cite{xue2018segan,you2022class,li2020shape,you2020unsupervised}, deep co-training \cite{qiao2018deep,zhou2019semi}, mean teacher schemes \cite{tarvainen2017mean,yu2019uncertainty,lai2022smoothed,lai2022sar,lai2021joint,he2021interpretable,you2022simcvd,you2023rethinking}, multi-task learning \cite{luo2020semi,kervadec2019curriculum,oliveira2022generalizable,you2022incremental,you2023implicit}, and contrastive learning \cite{chaitanya2020contrastive,wu2022exploring,you2021momentum,you2022mine,quan2022information,you2022bootstrapping}.

Contrastive learning (CL) has become a remarkable approach to enhance semi-supervised medical image segmentation performance without significantly increasing the amount of parameters and annotation costs \cite{chaitanya2020contrastive,wu2022exploring,you2022bootstrapping}. In real-world clinical scenarios, since the classes in medical images follow the Zipfian distribution \cite{zipf2013psycho}, the medical datasets usually show a long-tailed, even heavy-tailed class distribution, \ie, some minority (tail) classes involving significantly fewer pixel-level training instances than other majority (head) classes, as illustrated in Figure \ref{fig:distribution}. Such imbalanced scenarios are usually very challenging for CL methods to address, leading to noticeable performance drop \cite{li2020analyzing}. 

To address long-tail medical segmentation, our motivations come from the following two perspectives in CL training schemes \cite{chaitanya2020contrastive,you2022bootstrapping}: \ding{182} \textbf{Training objective} -- the main focus of existing approaches is on designing proper unsupervised contrastive loss in learning high-quality representations for long-tail medical segmentation. While extensively explored in the unlabeled portion of long-tail medical data, supervised CL has rarely been studied from empirical and theoretical perspectives, which will be one of the focuses in this work; 
\ding{183} \textbf{Temperature scheduler} -- the temperature parameter $\tau$, which controls the strength of attraction and repulsion forces in the contrastive loss \cite{chen2020improved,chen2020simple}, has been shown to play a crucial role in learning useful representations. It is affirmed that a large $\tau$ emphasizes anatomically meaningful group-wise patterns by group-level discrimination, whereas a small $\tau$ ensures a higher degree of pixel-level (instance) discrimination \cite{wang2021understanding,robinson2021can}. 
On the other hand, as shown in \cite{robinson2021can}, group-wise discrimination often results in reduced model's instance discrimination capabilities, where the model will be biased to ``easy'' features instead of ``hard'' features. 
It is thus unfavorable for long-tailed medical segmentation to blindly treat $\tau$ as a \textit{constant} hyperparameter, and a dynamic temperature parameter for CL is worth investigating.

In this paper, we introduce ACTION++, which further optimizes anatomically group-level and pixel-level representations for better head and tail class separations, on both labeled and unlabeled medical data. 
Specifically, we devise two strategies to improve overall segmentation quality by focusing on the two aforementioned perspectives: 
(1) we propose supervised adaptive anatomical contrastive learning (SAACL) for long-tail medical segmentation. To prevent the feature space from being biased toward the dominant head class, we first pre-compute the optimal locations of class centers uniformly distributed on the embedding space (\ie, off-line), and then perform online contrastive matching training by encouraging different class features to adaptively match these distinct and uniformly distributed class centers;
(2) we find that blindly adopting the \textit{constant} temperature $\tau$ in the contrastive loss can negatively impact the segmentation performance. 
Inspired by an average distance maximization perspective, we leverage a \textit{dynamic} $\tau$ via a simple cosine schedule, resulting in significant improvements in the learned representations. 
Both of these enable the model to learn a balanced feature space that has similar separability for both the majority (head) and minority (tail) classes, leading to better generalization in long-tail medical data. We evaluated our ACTION++ on the public ACDC and LA datasets \cite{acdc,la}. Extensive experimental
results show that our ACTION++ outperforms prior methods by a significant margin and sets the new state-of-the-art across two semi-supervised settings. We also theoretically show the superiority of our method in label efficiency (Appendix~\ref{sec:analysis}). Code is released at \href{https://github.com/charlesyou999648/ACTION}{here}.
\section{Method}
\label{section:method}

\begin{figure}[t]
\centering
\includegraphics[width=\linewidth]{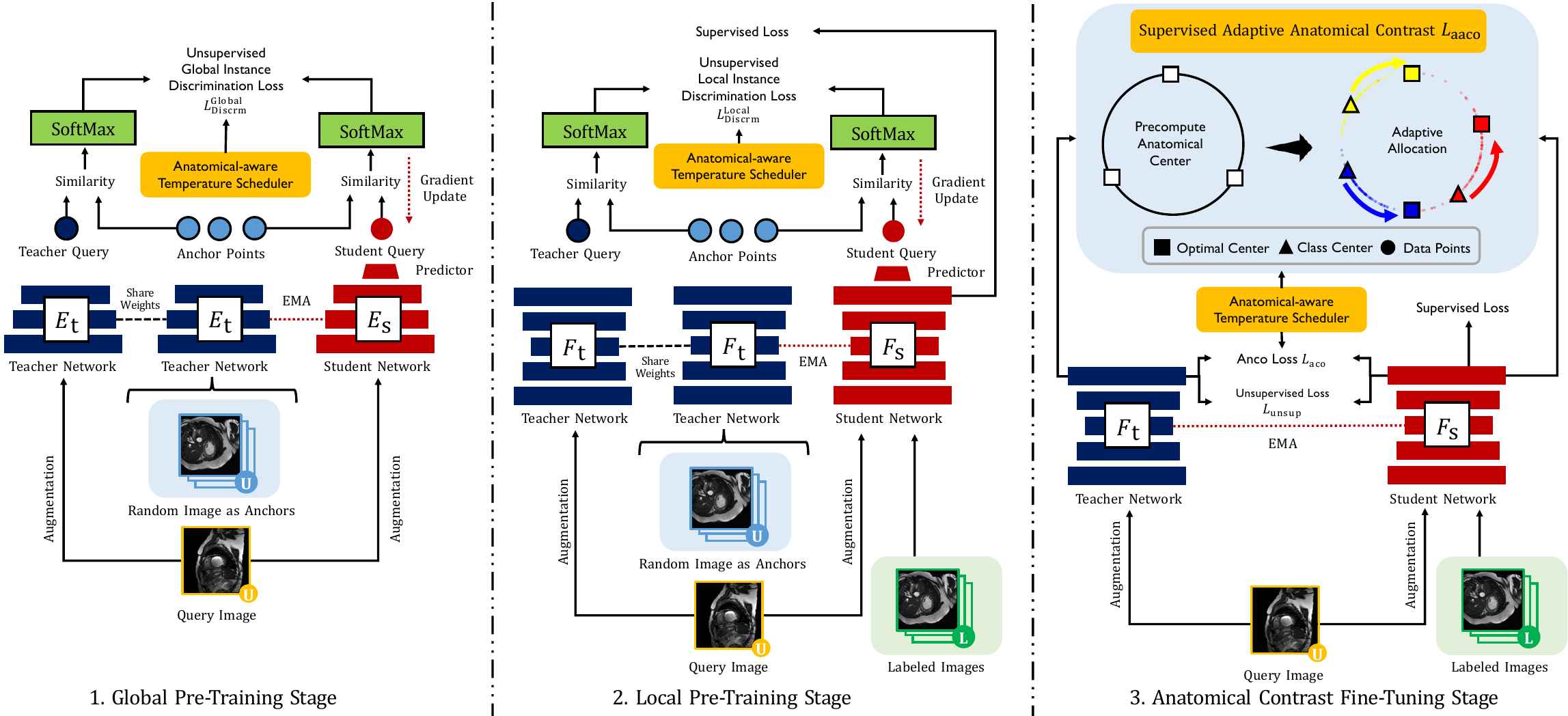}
\vspace{-20pt}
\caption{Overview of ACTION++: (1) global and local pre-training with proposed anatomical-aware temperature scheduler, (2) our proposed adaptive anatomical contrast fine-tuning, which first pre-computes the optimal locations of class centers uniformly distributed on the embedding space (\ie, off-line), and then performs online contrastive matching training by encouraging different class features to adaptively match these distinct and uniformly distributed class centers with respect to anatomical features.} 
\label{fig:model}
\vspace{-10pt}
\end{figure}

\subsection{Overview}
\myparagraph{Problem Statement}
Given a medical image dataset $(\boldsymbol{X}, \boldsymbol{Y})$, our goal is to train a segmentation model $\boldsymbol{F}$ that can provide accurate predictions that assign each pixel to their corresponding $K$-class segmentation labels.

\myparagraph{Setup}
Figure~\ref{fig:model} illustrates an overview of ACTION++. By default, we build this work upon ACTION pipeline \cite{you2022bootstrapping}, the state-of-the-art CL framework for semi-supervised medical image segmentation. The backbone model adopts the student-teacher framework that shares the same architecture, and the parameters of the teacher are the exponential moving average of the student’s parameters. Hereinafter, we adopt their model as our backbone and briefly summarize
its major components: (1) global contrastive distillation pre-training; (2) local contrastive distillation pre-training; and (3) anatomical contrast fine-tuning.

\myparagraph{Global and Local Pre-training}
\cite{you2022bootstrapping} first creates two types of anatomical views as follows: (1) {\em{augmented views}} - $\x^1$ and $\x^2$ are augmented from the unlabeled input scan with two separate data augmentation operators; (2) {\em{mined views}} - $n$ samples (\ie, $\x^3$) are randomly sampled from the unlabeled  portion with additional augmentation. The pairs $\left[\x^1,\x^2\right]$ are then processed by {student-teacher networks} $\left[F_{s},F_{t}\right]$ that share the same architecture and weight, and similarly, $\x^3$ is encoded by $F_{t}$. Their global latent features after the encoder $\boldsymbol{E}$ (\ie, $\left[\h^1,\h^2,\h^3\right]$) and local output features after decoder $\boldsymbol{D}$ (\ie, $\left[\f^1,\f^2,\f^3\right]$) are encoded by the two-layer nonlinear projectors, generating global and local embeddings $\vv_{g}$ and $\vv_{l}$. $\vv$ from $F_{s}$ are separately encoded by the non-linear predictor, producing $\w$ in both global and local manners\,\footnote{For simplicity, we omit details of local instance discrimination in the following.}. Third, the relational similarities between {{augmented}} and {{mined}} views are processed by SoftMax function as follows: 
$\uu_s = \text{log} \frac{\text{exp}\big(\text{sim}\big({\w^1}, {\vv^{3}}\big)/\tau_s\big)}{\sum_{n=1}^{N} \text{exp}\big(\text{sim}\big({\w^1}, {\vv^{3}_{n}}\big)/\tau_s\big)},\,
\uu_t = \text{log} \frac{\text{exp}\big(\text{sim}\big({\w^2}, {\vv^{3}}\big)/\tau_t\big)}{\sum_{n=1}^{N} \text{exp}\big(\text{sim}\big({\w^2}, {\vv^{3}_{n}}\big)/\tau_t\big)},$
where $\tau_s$ and $\tau_t$ are two temperature parameters. Finally, we minimize the unsupervised instance discrimination loss (\ie, Kullback-Leibler divergence $\mathcal{KL}$) as:
\vspace{-0.2cm}
\begin{equation}
\mathcal{L}_{\text{inst}} = \mathcal{KL}(\uu_s || \uu_t).
\label{equation:pcl}
\end{equation}
We formally summarize the pretraining objective as the equal combination of the global and local $\mathcal{L}_{\text{inst}}$, and supervised segmentation loss $\mathcal{L}_\text{sup}$ (\ie, equal combination of Dice loss and cross-entropy loss).

\myparagraph{Anatomical Contrast Fine-tuning} 
The underlying motivation for the fine-tuning stage is that it reduces the vulnerability of the pre-trained model to long-tailed unlabeled data. To mitigate the problem, \cite{you2022bootstrapping} proposed to fine-tune the model by anatomical contrast. First, the additional representation head $\boldsymbol{\varphi}$ is used to provide dense representations with the same size as the input scans. Then, \cite{you2022bootstrapping} explore pulling queries $\rr_q\!\in\! \mathcal{R}$ to be similar to the positive keys $\rr_k^{+}\!\in\! \mathcal{R}$, and push apart the negative keys $\rr_k^{-}\!\in\! \mathcal{R}$. The AnCo loss is defined as follows:
\begin{equation}
  \label{loss:anco}
   \mathcal{L}_\text{anco} = \sum_{c\in \mathcal{C}} \sum_{\rr_q \sim \mathcal{R}^c_q} -\log \frac{\exp(\rr_q \cdot \rr_k^{c, +} / \tau_{an})}{\exp(\rr_q \cdot \rr_k^{c, +}/ \tau_{an}) + \sum_{\rr_k^{-}\sim \mathcal{R}^c_k} \exp(\rr_q \cdot \rr_k^{-}/ \tau_{an})},
\end{equation}
where $\mathcal{C}$ denotes a set of all available classes in the current mini-batch, and $\tau_{an}$ is a temperature hyperparameter. For class $c$, we select a query representation set $\mathcal{R}_q^c$, a negative key representation set $\mathcal{R}_k^c$ whose labels are not in class $c$, and the positive key $\rr_k^{c, +}$ which is the $c$-class mean representation. Given $\mathcal{P}$ is a set including all pixel coordinates with the same size as $R$, these queries and keys can be defined as: $\mathcal{R}_q^c\!=\!\!\bigcup_{[i, j]\in \mathcal{A}}\!\!\mathbbm{1}(\y_{[i,j]}\!=\!c)\, \rr_{[i,j]},\,  \mathcal{R}_k^c\! =\!\!\bigcup_{[i, j]\in \mathcal{A}}\!\!\mathbbm{1}(\y_{[i,j]}\!\neq\!c)\, \rr_{[i,j]},\,  \rr_k^{c, +}\!\! =\!\frac{1}{| \mathcal{R}_q^c |}\sum_{\rr_q \in \mathcal{R}_q^c} \rr_q\,$. We formally summarize the fine-tuning objective as the equal combination of unsupervised $\mathcal{L}_\text{anco}$, unsupervised cross-entropy loss $\mathcal{L}_\text{unsup}$, and  supervised segmentation loss $\mathcal{L}_\text{sup}$. For more details, we refer the reader to \cite{you2022bootstrapping}.

\subsection{Supervised Adaptive Anatomical Contrastive Learning}\label{sec: saacl}
The general efficacy of anatomical contrast on long-tail unlabeled data has previously been demonstrated by the authors of \cite{you2022bootstrapping}. However, taking a closer look, we observe that the well-trained $\boldsymbol{F}$ shows a downward trend in performance, which often fails to classify tail classes on labeled data, especially when the data shows long-tailed class distributions. This indicates that such well-trained $\boldsymbol{F}$ is required to improve the segmentation capabilities in long-tailed labeled data. 
To this end, inspired by \cite{li2022targeted} tailored for the image classification tasks, we introduce supervised adaptive anatomical contrastive learning (SAACL), a training framework for generating well-separated and uniformly distributed latent feature representations for both the head and tail classes. 
It consists of three main steps, which we describe in the following.

\paragraph{Anatomical Center Pre-computation} We first pre-compute the anatomical class centers in latent representation space.  
The optimal class centers are chosen as $K$ positions from the unit sphere $\mathbb{S}^{d-1} = \{ v \in \mathbb{R}^{d}: \ \|v\|_2 = 1 \}$ in the $d$-dimensional space. 
To encourage good separability and uniformity, we compute the class centers $\{\bm{\psi}_c\}_{c=1}^K$ by minimizing the following uniformity loss $\mathcal{L}_\text{unif}$:
\begin{equation}
  \label{loss:unif}
    \mathcal{L}_\text{unif} (\{\bm{\psi}_c\}_{c=1}^K) = \sum_{c=1}^K \log \left( \sum_{c'=1}^K \exp({ \bm{\psi}_c \cdot \bm{\psi}_{c'} /\tau}) \right).
\end{equation}
In our implementation, we use gradient descent to search for the optimal class centers constrained to the unit sphere $\mathbb{S}^{d-1}$, which are denoted by $\{\bm{\psi}^\star_c\}_{c=1}^K$. 
Furthermore, the latent dimension $d$ is a hyper-parameter, which we set such that $d \gg K$ to ensure the solution found by gradient descent indeed maximizes the minimum distance between any two class centers \cite{graf2021dissecting}. 
It is also known that any analytical minimizers of  Eqn. \ref{loss:unif} form a perfectly regular $K$-vertex inscribed simplex of the sphere $\mathbb{S}^{d-1}$ \cite{graf2021dissecting}. 
We emphasize that this first step of pre-computation of class centers is completely off-line as it does not require any training data. 

\paragraph{Adaptive Allocation} 
As the second step, we explore adaptively allocating these centers among classes. 
This is a combinatorial optimization problem and an exhaustive search of all choices would be computationally prohibited.
Therefore, we draw intuition from the empirical mean in the K-means algorithm and adopt an adaptive allocation scheme to iteratively search for the optimal allocation during training. 
Specifically, consider a batch $\mathcal{B}=\{\mathcal{B}_1, \cdots, \mathcal{B}_K\}$ where $\mathcal{B}_c$ denotes a set of samples in a batch with class label $c$, for $c = 1,\cdots, K$. 
Define $\overline{\bm{\phi}}_c (\mathcal{B}) = \sum_{i \in \mathcal{B}_c} \bm{\phi}_i / \|\sum_{i \in \mathcal{B}_c} \bm{\phi}_i \|_2$ be the empirical mean of class $c$ in current batch, where $\bm{\phi}_i$ is the feature embedding of sample $i$.
We compute assignment $\pi$ by minimizing the distance between pre-computed class centers and the empirical means:
\begin{equation}
    \pi^\star = \arg\min_{\pi} \sum_{c=1}^K \|\bm{\psi}_{\pi(c)}^\star - \overline{\bm{\phi}}_c \|_2. 
\end{equation}
In implementation, the empirical mean is updated using moving average. That is, for iteration $t$, we first compute the empirical mean $\overline{\bm{\phi}}_c (\mathcal{B})$ for batch $\mathcal{B}$ as  described above, and then update by $\overline{\bm{\phi}}_c \leftarrow (1-\eta) \overline{\bm{\phi}}_c + \eta \overline{\bm{\phi}}_c (\mathcal{B})$. 

\paragraph{Adaptive Anatomical Contrast} 
Finally, the allocated class centers are well-separated and should maintain the semantic relation between classes. 
To utilize these optimal class centers, we want to induce the feature representation of samples from each class to cluster around the corresponding pre-computed class center.
To this end, we adopt a supervised contrastive loss for the label portion of the data. 
Specifically, given a batch of pixel-feature-label tuples $\{(\omega_i, \bm{\phi}_i, y_i)\}_{i=1}^n$ where $\omega_i$ is the i-th pixel in the batch, $\bm{\phi}_i$ is the feature of the pixel and $y_i$ is its label, we define supervised \underline{\tf{a}}daptive \underline{\tf{a}}natomical \underline{\tf{co}}ntrastive loss for pixel $i$ as:
\begin{equation}
  \label{loss:aaco}
    \mathcal{L}_{\text{aaco}} = \frac{-1}{n} \sum_{i=1}^n \left( \sum_{\bm{\phi}_i^+} \log \frac{\exp(\bm{\phi}_i \cdot \bm{\phi}_i^+/\tau_{sa})}{\sum_{\bm{\phi}_j} \exp(\bm{\phi}_i \cdot \bm{\phi}_j /\tau_{sa})} + \lambda_{a} \log \frac{\exp(\bm{\phi}_i \cdot \bm{\nu}_i /\tau_{sa})}{\sum_{\bm{\phi}_j} \exp(\bm{\phi}_i \cdot \bm{\phi}_j /\tau_{sa})} \right),
\end{equation}
where $\bm{\nu}_i = \bm{\psi}_{\pi^\star(y_i)}^\star$ is the pre-computed center of class $y_i$. 
The first term in Eqn. \ref{loss:aaco} is supervised contrastive loss, where the summation over $\bm{\phi}_i^+$ refers to the uniformly sampled positive examples from pixels in batch with label equal to $y_i$. 
The summation over $\bm{\phi}_j$ refers to all features in the batch excluding $\bm{\phi_i}$. 
The second term is contrastive loss with the positive example being the pre-computed optimal class center.

\subsection{Anatomical-aware Temperature Scheduler (ATS)}
Training with a varying $\tau$ induces a more isotropic representation space, wherein the model learns both group-wise and instance-specific features \cite{kukleva2023temperature}. 
To this end, we are inspired to use an anatomical-aware temperature scheduler in both the supervised and the unsupervised contrastive losses, where the temperature parameter $\tau$ evolves within the range $[\tau^-, \tau^+]$ for $\tau^+ > \tau^-$.
Specifically, for iteration $t = 1,\cdots, T$ with $T$ being the total number of iterations, we set $\tau_t$ as:
\begin{equation}
    \tau_t = \tau^- + 0.5 (1+\cos(2\pi t/ T)) (\tau^+ - \tau^-). 
\end{equation}
\section{Experiments}
\label{section:exp}
\myparagraph{Experimental Setup}
We evaluate ACTION++ on two benchmark datasets: the LA dataset \cite{la} and the ACDC dataset \cite{acdc}. The LA dataset consists of 100 gadolinium-enhanced MRI scans, with the fixed split \cite{wu2022exploring} using 80 and 20 scans for training and validation.
The ACDC dataset consists of 200 cardiac cine MRI scans from 100 patients including three segmentation classes, \ie, left ventricle (LV), myocardium (Myo), and right ventricle (RV), with the fixed split\footnote{\url{https://github.com/HiLab-git/SSL4MIS/tree/master/data/ACDC}} using 70, 10, and 20 patients' scans for training, validation, and testing. 
For all our experiments, we follow the identical setting in \cite{yu2019uncertainty,luo2020semi,wu2021semi,wu2022exploring}, and perform evaluations under two label settings (\ie, 5\% and 10\% ) for both datasets.

\begin{table*}[t]
	\begin{center}
	\caption{Quantitative comparison (DSC{[}\%{]}/ASD{[}voxel{]}) for LA under two unlabeled settings (5\% or\! 10\%). All experiments are conducted as \cite{yu2019uncertainty,li2020shape,luo2020semi,luo2021efficient,wu2021semi,wu2022exploring,you2022bootstrapping} in the identical setting for fair comparisons. The best results are indicated in \tf{bold}. VNet-F (fully-supervided) and VNet-L (semi-supervided) are considered as the upper bound and the lower bound for the performance comparison.}
	\vspace{-6pt}
	\label{table:la_main}
    \begin{adjustbox}{width=0.7\linewidth}
	\begin{tabular}{ccccc}
		\toprule
		& \multicolumn{2}{c}{4 Labeled (5\%)} & \multicolumn{2}{c}{8 Labeled (10\%)} \\
        \cmidrule(r){2-3} \cmidrule(r){4-5}
		{Method}
		            & {DSC{[}\%{]}$\uparrow$}  
		            & {ASD{[}voxel{]}$\downarrow$}
		            & {DSC{[}\%{]}$\uparrow$}  
		            & {ASD{[}voxel{]}$\downarrow$}
		            \\ \midrule
		VNet-F \cite{milletari2016v}
		            & {91.5}
                    & {1.51}
		            & {91.5}
                    & {1.51}
                    \\
		VNet-L        
		            & {52.6}
                    & {9.87}
                    & {82.7}
                    & {3.26}
                    \\\midrule 
	    UAMT \cite{yu2019uncertainty} 
                    & {82.3}
                    & {3.82}
                    & {87.8}
                    & {2.12}
                    \\ 
	    SASSNet \cite{li2020shape} 
                    & {81.6}
                    & {3.58}
                    & {87.5}
                    & {2.59}
                    \\
	    DTC \cite{luo2020semi} 
                    & {81.3}
                    & {2.70}
                    & {87.5}
                    & {2.36}
                    \\ 
	    URPC \cite{luo2021efficient} 
                    & {82.5}
                    & {3.65}
                    & {86.9}
                    & {2.28}
                    \\ 
	    MC-Net \cite{wu2021semi} 
                    & {83.6}
                    & {2.70}
                    & {87.6}
                    & {1.82}
                    \\ 
	    SS-Net \cite{wu2022exploring} 
                    & {86.3}
                    & {2.31}
                    & {88.6}
                    & {1.90}
                    \\
        ACTION \cite{you2022bootstrapping}
                    & {86.6}
                    & {2.24}
                    & {88.7}
                    & {2.10}
                    \\ 
        \gr \cb ACTION++ (ours)
                    & \tf{87.8}
                    & \tf{2.09}
                    & \tf{89.9}
                    & \tf{1.74}
                    \\ 
		              \bottomrule
	\end{tabular}
    \end{adjustbox}
    \end{center}
    \vspace{-30pt}
\end{table*}

\myparagraph{Implementation Details}
We use an SGD optimizer  for all experiments with a learning rate of 1e-2, a momentum of $0.9$, and a weight decay of $0.0001$. Following \cite{yu2019uncertainty,luo2020semi,wu2021semi,wu2022exploring} on both datasets, all inputs were normalized as zero mean and unit variance. The data augmentations are rotation and flip operations.
Our work is built on ACTION \cite{you2022bootstrapping}, thus we follow the identical model setting except for temperature parameters because they are of direct interest to us. 
For the sake of completeness, we refer the reader to \cite{you2022bootstrapping} for more details. 
We set $\lambda_{a}$, $d$ as 0.2, 128, and regarding all $\tau$, we use $\tau^{+}$=1.0 and $\tau^{-}$=0.1 if not stated otherwise.
On ACDC, we use the U-Net model \cite{ronneberger2015u} as the backbone with a 2D patch size of $256\times256$ and batch size of 8.
For pre-training, the networks are trained for 10K iterations; for fine-tuning, 20K iterations.
On LA, we use the V-Net \cite{milletari2016v} as the backbone. For training, we randomly crop $112\times112\times80$ patches and the batch size is 2. For pre-training, the networks are trained for 5K iterations. For fine-tuning, the networks are for 15K iterations. For testing, we adopt a sliding window strategy with a fixed stride ($18\times18\times4$). All experiments are conducted in the same environments with fixed random seeds (Hardware: Single NVIDIA GeForce RTX 3090 GPU; Software: PyTorch 1.10.2+cu113, and Python 3.8.11).

\myparagraph{Main Results}
We compare our ACTION++ with current state-of-the-art SSL methods, including UAMT \cite{yu2019uncertainty}, SASSNet \cite{li2020shape}, DTC \cite{luo2020semi}, URPC \cite{luo2021efficient}, MC-Net \cite{wu2021semi}, SS-Net \cite{wu2022exploring}, and ACTION \cite{you2022bootstrapping}, and the supervised counterparts (UNet \cite{ronneberger2015u}/VNet \cite{milletari2016v}) trained with Full/Limited supervisions -- using their released code. To evaluate 3D segmentation ability, we use Dice coefficient (DSC) and Average Surface Distance (ASD). Table \ref{table:acdc_main} and Table \ref{table:la_main} display the results on the public ACDC and LA datasets for the two labeled settings, respectively. We next discuss our main findings as follows. (1) \textbf{LA}: As shown in Table \ref{table:la_main}, our method generally presents better performance than the prior SSL methods under all settings. Fig. \ref{fig:vis_la} (Appendix) also shows that our model consistently outperforms all other competitors, especially in the boundary region;
(2) \textbf{ACDC}: As Table \ref{table:acdc_main} shows, ACTION++ achieves the best segmentation performance in terms of Dice and ASD, consistently outperforming the previous SSL methods across two labeled settings. In Fig. \ref{fig:vis_acdc} (Appendix), we can observe that ACTION++ can yield the segmentation boundaries accurately, even for very challenging regions (\ie, RV and Myo). This suggests that ACTION++ is inherently better at long-tailed learning, in addition to being a better segmentation model in general.
\begin{table*}[t]
	\begin{center}
	\caption{Quantitative comparison (DSC{[}\%{]}/ASD{[}voxel{]}) for ACDC under two unlabeled settings (5\% or 10\%). All experiments are conducted as \cite{yu2019uncertainty,li2020shape,luo2020semi,luo2021efficient,wu2021semi,wu2022exploring,you2022bootstrapping} in the identical setting for fair comparisons. The best results are indicated in \tf{bold}.}
	\vspace{-16pt}
	\label{table:acdc_main}
    \begin{adjustbox}{width=\linewidth}
	\begin{tabular}{ccccccccc}
		\toprule
		& & \multicolumn{3}{c}{3 Labeled (5\%)} & & \multicolumn{3}{c}{7 Labeled (10\%)} \\
        \cmidrule(r){3-5} \cmidrule(r){7-9}
		{Method}
		            & {Average}
		            & {RV}  
		            & {Myo}
		            & {LV}
		            & {Average}
		            & {RV}  
		            & {Myo}
		            & {LV} 
		            \\ \midrule
		UNet-F \cite{ronneberger2015u}
		            & {91.5}/{0.996} 
		            & {90.5}/{0.606}
                    & {88.8}/{0.941}
                    & {94.4}/{1.44}
		            & {91.5}/{0.996} 
		            & {90.5}/{0.606}
                    & {88.8}/{0.941}
                    & {94.4}/{1.44}
                    \\ 
		UNet-L        
                    & {51.7}/{13.1} 
		            & {36.9}/{30.1}
                    & {54.9}/{4.27}
                    & {63.4}/{5.11}
                    & {79.5}/{2.73}
		            & {65.9}/{0.892}
                    & {82.9}/{2.70}
                    & {89.6}/{4.60}
                    \\\midrule 
	    UAMT \cite{yu2019uncertainty} 
                    & {48.3}/{9.14}
                    & {37.6}/{18.9}
                    & {50.1}/{4.27}
                    & {57.3}/{4.17}
                    & {81.8}/{4.04}
                    & {79.9}/{2.73}
                    & {80.1}/{3.32}
                    & {85.4}/{6.07}
                    \\ 
	    SASSNet \cite{li2020shape} 
                    & {57.8}/{6.36}
                    & {47.9}/{11.7}
                    & {59.7}/{4.51}
                    & {65.8}/{2.87}
                    & {84.7}/{1.83}
                    & {81.8}/{0.769}
                    & {82.9}/{1.73}
                    & {89.4}/{2.99}
                    \\
	    URPC \cite{luo2021efficient} 
                    & {58.9}/{8.14}
                    & {50.1}/{12.6}
                    & {60.8}/{4.10}
                    & {65.8}/{7.71}
                    & {83.1}/{1.68}
                    & {77.0}/{0.742}
                    & {82.2}/{0.505}
                    & {90.1}/{3.79}
                    \\ 
	    DTC \cite{luo2020semi} 
                    & {56.9}/{7.59}
                    & {35.1}/{9.17}
                    & {62.9}/{6.01}
                    & {72.7}/{7.59}
                    & {84.3}/{4.04}
                    & {83.8}/{3.72}
                    & {83.5}/{4.63}
                    & {85.6}/{3.77}
                    \\ 
	    MC-Net \cite{wu2021semi} 
                    & {62.8}/{2.59}
                    & {52.7}/{5.14}
                    & {62.6}/{0.807}
                    & {73.1}/{1.81}
                    & {86.5}/{1.89}
                    & {85.1}/{0.745}
                    & {84.0}/{2.12}
                    & {90.3}/{2.81}
                    \\ 
	    SS-Net \cite{wu2022exploring} 
                    & {65.8}/{2.28}
                    & {57.5}/{3.91}
                    & {65.7}/{2.02}
                    & {74.2}/{0.896}
                    & {86.8}/{1.40}
                    & {85.4}/{1.19}
                    & {84.3}/{1.44}
                    & {90.6}/{1.57}
                    \\
        ACTION \cite{you2022bootstrapping}
                    & {87.5}/{1.12}
                    & {85.4}/{0.915}
                    & {85.8}/{0.784}
                    & {91.2}/{1.66}
                    & {89.7}/{0.736}
                    & {89.8}/{0.589}
                    & {86.7}/{0.813}
                    & {92.7}/{0.804}
                    \\ 
        \gr \cb ACTION++ (ours)
                    & \tf{88.5}/\tf{0.723}
                    & \tf{86.9}/\tf{0.662}
                    & \tf{86.8}/\tf{0.689}
                    & \tf{91.9}/\tf{0.818}
                    & \tf{90.4}/\tf{0.592}
                    & \tf{90.5}/\tf{0.448}
                    & \tf{87.5}/\tf{0.628}
                    & \tf{93.1}/\tf{0.700}
                    \\ 
		              \bottomrule
	\end{tabular}
    \end{adjustbox}
    \end{center}
    \vspace{-20pt}
\end{table*}

\myparagraph{Ablation Study}
We first perform ablation studies on LA with 10\% label ratio to evaluate the importance of different components. Table \ref{table:ablation_aaco} shows the effectiveness of supervised adaptive
anatomical contrastive learning (SAACL). 
Table \ref{table:ablation_cos} (Appendix) indicates that using anatomical-aware temperature scheduler (ATS) and SAACL yield better performance in both pre-training and fine-tuning stages. We then theoretically show the superiority of our method in Appendix~\ref{sec:analysis}.

Finally, we conduct experiments to study the effects of cosine boundaries, cosine period, different methods of varying $\tau$, and $\lambda_a$ in Table \ref{table:ablation_ats}, Table \ref{table:ablation_scheduler} (Appendix), respectively. Empirically, we find that using our settings (\ie, $\tau^{-}\!=\!0.1$, $\tau^{+}\!=\!1.0$, $T$/\#iterations=1.0, cosine scheduler, $\lambda_{a}=0.2$) attains optimal performance.

\begin{table}
\vspace{-5mm}
\parbox{.58\linewidth}{
\caption{Ablation studies of Supervised Adaptive Anatomical Contrast (SAACL).}
\vspace{-8pt}
\label{table:ablation_aaco}
\resizebox{\linewidth}{!}
{
\begin{tabular}{@{\hskip 1mm}lcccccc@{\hskip 1mm}}
\toprule
Method  && {DSC{[}\%{]}$\uparrow$} && ASD{[}voxel{]$\downarrow$}\\ 
\midrule
KCL \cite{kang2021exploring}  && 88.4 && 2.19 \\
CB-KCL \cite{kang2019decoupling}  && 86.9 && 2.47 \\
SAACL (Ours) && \tf{89.9} && \tf{1.74} \\
\midrule
SAACL (random assign)  && 88.0 && 2.79 \\
SAACL (adaptive allocation) && \tf{89.9} && \tf{1.74} \\
\bottomrule
\end{tabular}
}
}
\hfill
\parbox{.38\linewidth}{
\vspace{-0.01mm}
\caption{Effect of cosine boundaries in with the largest difference between $\tau^{-}$ and $\tau^{+}$.}
\label{table:ablation_cos}
\vspace{-8pt}
\resizebox{\linewidth}{!}
{
\begin{tabular}{@{\hskip 1mm}l|ccccc@{\hskip 1mm}}
\toprule
\backslashbox{$\tau^{-}$}{$\tau^+$}  & 0.2 & 0.3 & 0.4 & 0.5 & 1.0 \\ 
\midrule
0.07 & 84.1 & 85.0 & 86.9 & 87.9 & 89.7 \\
0.1 & 84.5 & 85.9 & 87.1 & 88.3 & \tf{89.9} \\
0.2 & 84.2 & 84.4 & 85.8 & 87.1 & 87.6 \\
\bottomrule
\end{tabular}
}
}
\vspace{-5mm}
\end{table}

\section{Conclusion}
In this paper, we proposed ACTION++, an improved contrastive learning framework with adaptive anatomical contrast for semi-supervised medical segmentation. Our work is inspired by two intriguing observations that, besides the unlabeled data, the class imbalance issue exists in the labeled portion of medical data and the effectiveness of temperature schedules for contrastive learning on long-tailed medical data. Extensive experiments and ablations demonstrated that our model consistently achieved superior performance compared to the prior semi-supervised medical image segmentation methods under different label ratios.
Our theoretical analysis also revealed the robustness of our method in label efficiency.
In future, we will validate CT/MRI datasets with more foreground labels and try t-SNE.

%
%
\bibliographystyle{splncs04}
\bibliography{mybibliography}

\clearpage
\appendix 
\section*{Appendix}

\begin{figure}[ht]
\vspace{-10mm}
\centering
\includegraphics[width=\linewidth]{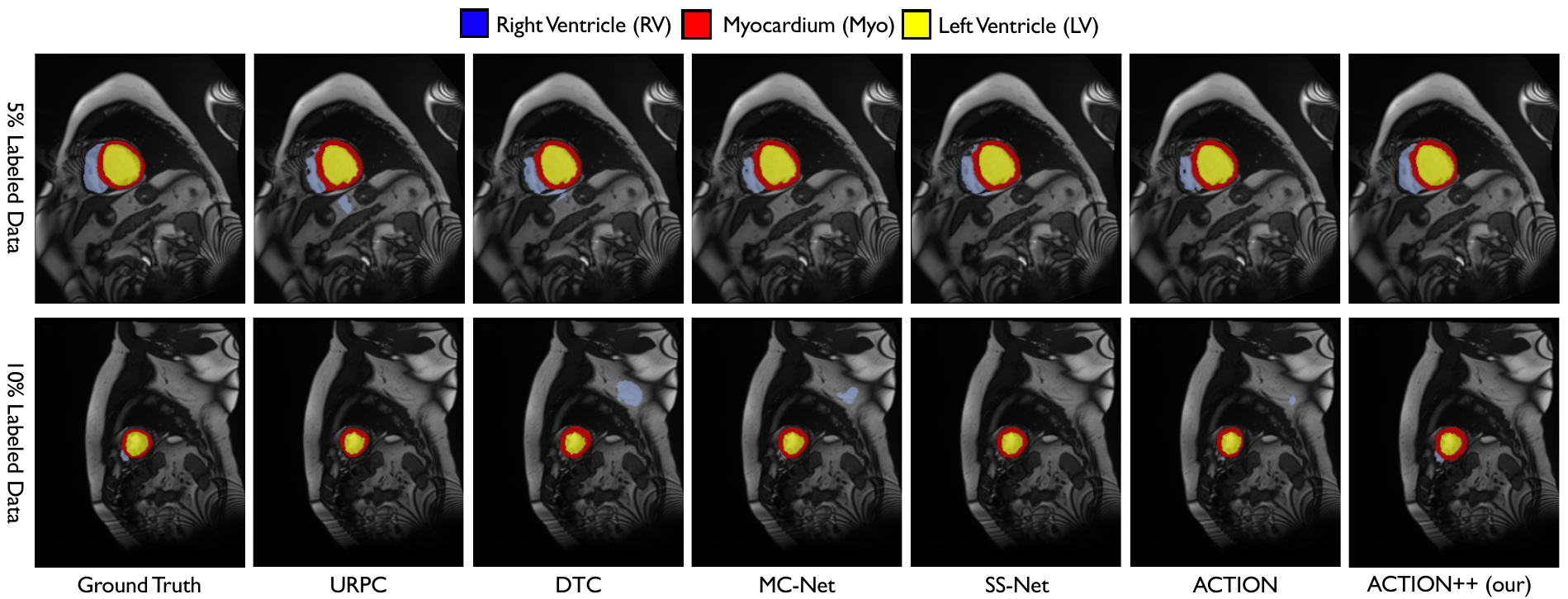}
\vspace{-20pt}
\caption{Visualization results on ACDC with 5\% and 10\% labeled data. ACTION++ consistently outputs more accurate predictions, especially for small regions.} 
\label{fig:vis_acdc}
\vspace{-15pt}
\end{figure}

\begin{figure}[h]
\vspace{-10mm}
\centering
\includegraphics[width=\linewidth]{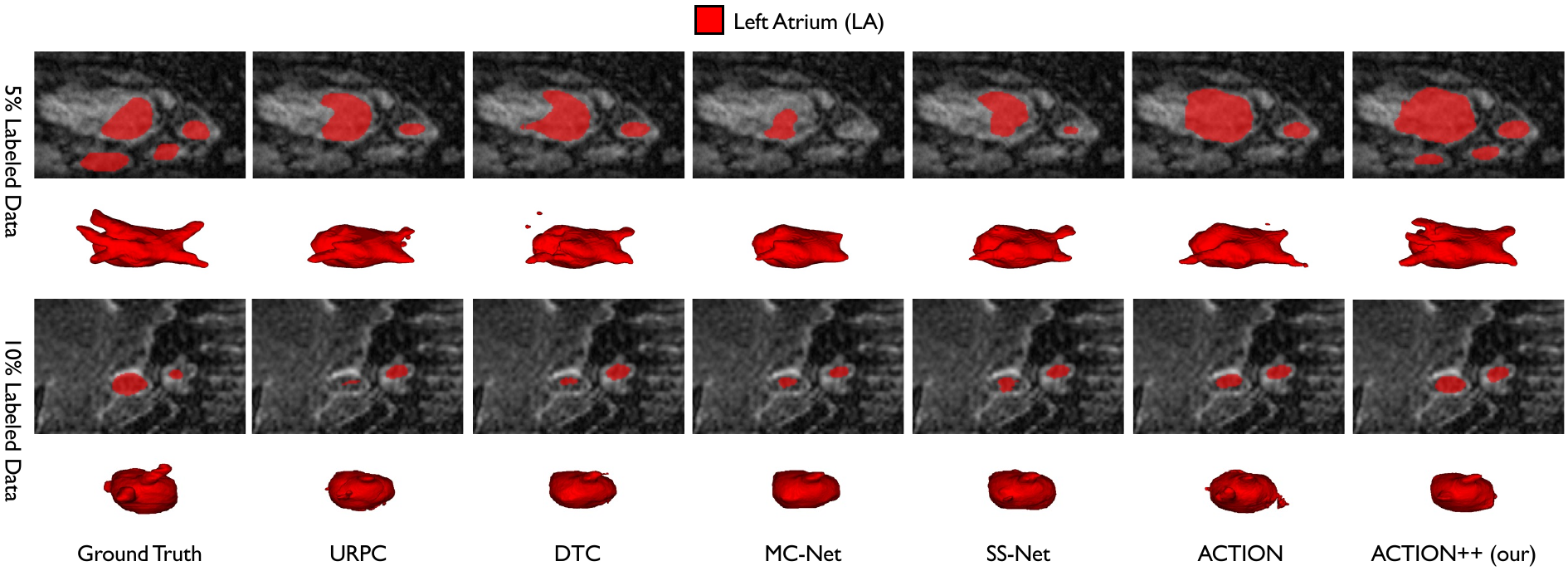}
\vspace{-20pt}
\caption{Visualization results on LA with 5\% and 10\% labeled data. ACTION++ consistently achieves more sharper and accurate object boundaries.} 
\label{fig:vis_la}
\vspace{-15pt}
\end{figure}

\begin{table}
\vspace{-3mm}
\parbox{.43\linewidth}{
\caption{Ablation studies of different components (\ie, ATS and SAACL).}
\vspace{-8pt}
\label{table:ablation_ats}
\resizebox{\linewidth}{!}
{
\begin{tabular}{@{\hskip 1mm}lcccccc@{\hskip 1mm}}
\toprule
Method  && {DSC{[}\%{]}$\uparrow$} && ASD{[}voxel{]$\downarrow$}\\ 
\midrule
pre-training w/o ATS && 86.2 && 2.69 \\
pre-training w/ ATS && 88.1 && 2.44 \\
\midrule
fine-tuning w/o SAACL/ATS  && 89.0 && 2.06 \\
fine-tuning only w/ ATS && {89.3} && {1.98} \\
fine-tuning only w/ SAACL && {89.5} && {1.96} \\
fine-tuning w/ SAACL/ATS && \tf{89.9} && \tf{1.74} \\
\bottomrule
\end{tabular}
}
}
\hfill
\parbox{.535\linewidth}{
\vspace{-0.2mm}
\caption{Effect of cosine period, different methods of varying $\tau$, and $\lambda_a$.}\label{table:ablation_scheduler}
\vspace{-8pt}
\resizebox{\linewidth}{!}
{
\begin{tabular}{@{\hskip 1mm}cccccc@{\hskip 3mm}ccccccc@{\hskip 1mm}}
\toprule
$T$/\#iterations  && DSC && Method && DSC && $\lambda_a$ && DSC \\ 
\cmidrule(l{0mm}r{0.5mm}){1-3} \cmidrule(l{0.3mm}r{0mm}){5-7} \cmidrule(l{0.3mm}r{0mm}){9-11}
no/fixed $\tau$  && 89.5 && fixed && 89.5 && 0.05 && 88.5 \\
0.1 && 89.8 && step && 89.4 && 0.1  && 89.3 \\
0.2 && 89.1 && rand && 88.9 && 0.2  && \textbf{89.9} \\
0.5 && 89.2 && oscil. && 89.2 && 0.5  && 89.7 \\
1.0 && \textbf{89.9} && cos && \textbf{89.9} && 1.0  && 89.1 \\
2.0 && 89.7 && - && - && 10  && 87.9 \\
\bottomrule
\end{tabular}
}
}
\vspace{-10mm}
\end{table}

\section{Theoretical Analysis}\label{sec:analysis}

In this section, we discuss the performance guarantee of the proposed SAACL. 
For abstraction, we denote an image and its corresponding segmentation map as $\mathbf{x} = \{\omega_p\}_p$, $\mathbf{y}=\{y_p\}_p$, where $\omega_p$ is a pixel. 
We also denote the feature generator as $f$, such that $f(\omega_p; \mathbf{x}) = \bm{\phi}_p$ for any pixel $p$. 
Recent work \cite{huang2021towards} has shown that, to evaluate the performance of the representations learned via contrastive learning (CL), it suffices to consider a simplified nearest neighbour (NN) classifier\footnote{This is because an NN classifier is a special case of a linear classifier, which can be approximated by a neural network. See Sec.~2 of \cite{huang2021towards}.} $g_f(\omega_p; \mathbf{x}) = \argmin_{c \in [K]} \| f(\omega_p; \mathbf{x}) - \psi^\star_c \|_2$, where $\psi^\star_c$ denotes the center of class $c$ in the latent representation space.
To this end, we focus on the error rate of $g_f$ defined as $\mathcal{E}(g_f) = \sum_{c=1}^K \mathbb{P}[g_f(\omega_p;\mathbf{x})\neq c, \forall \omega_p \in Cla_c]$, where $\omega_p \in Cla_c$ refers to pixels in class $c$. 
Note that each class $c$, regardless of being head or tail class, has equal weight in the definition of $\mathcal{E}(g_f)$, indicating that a small $\mathcal{E}(g_f)$ implies good long-tail segmentation performance.

We now demonstrate that SAACL helps achieve a small error $\mathcal{E}(g_f)$.
The success of contrastive learning mainly depends on two aspects: positive alignment and class divergence~\cite{huang2021towards}. 
Specifically, the positive alignment is defined as follows:
\begin{align}\label{eq: positive alignment}
    A = \sqrt{\mathbb{E}_{\mathbf{x},\tilde{\mathbf{x}}} \mathbb{E}_{c \in [K]} \mathbb{E}_{\omega_p \in Cla_c} [\| f(\omega_p; \mathbf{x}) - f(\omega_p; \tilde{\mathbf{x}})\|^2]}, 
\end{align} 
where $\mathbf{x}$ and $\tilde{\mathbf{x}}$ are two augmentations from the same input sample (\ie, positive sample pairs). 
The class divergence is defined as $D = \max_{c \neq c'}  \overline{\bm{\phi}}_c \cdot \overline{\bm{\phi}}_{c'}  $, which computes the distances between class centers. 
The following theorem discloses the link between the error rate and the alignment $A$ and class divergence $D$. 
\begin{theorem}[\cite{huang2021towards}]\label{thm:main}
    There exist some constant $\rho(\sigma, \delta, \epsilon)$ and $\Delta$ whose value depends on the data augmentation method and Lipschitzness of the model $f$. 
    Let $\zeta(\sigma, \delta, \epsilon)= r^2 [1-\rho(\sigma, \delta, \epsilon) - \sqrt{2\rho(\sigma,\delta, \epsilon)}-\Delta/2]$.
    If for any class $c, c' \in [K]$, it holds that $\overline{\bm{\phi}}_{c}\cdot \overline{\bm{\phi}}_{c'} \leq \zeta(\sigma, \delta, \epsilon)$, then $\mathcal{E} (g_f) \leq 1 - \sigma + \mathcal{O}(1/\epsilon) A $. 
\end{theorem}Due to space limit, please refer to Theorem~1 in \cite{huang2021towards} for the detailed mathematical form of $\rho(\sigma, \delta, \epsilon)$, $\Delta$ and the problem-related parameters $\sigma$, $\delta$ and $\epsilon$.  
For our purpose, we observe that: (1) good positive alignment (small $A$) directly indicates low error according to the error upper bound; (2) a large class divergence (small $D$) can help satisfy the condition on $\overline{\bm{\phi}}_{c}\cdot \overline{\bm{\phi}}_{c'}$. Therefore, both $A$ are $D$ are crucial to improving the representation
learning.

From \eqref{loss:aaco}, both the alignment and the diversity are captured by the objective $\mathcal{L}_{\text{aaco}}$.
We rewrite \eqref{loss:aaco} as $\mathcal{L}_{\text{aaco}} = \sum_{i=1}^n(\mathcal{L}_{i,1}+\lambda_a \mathcal{L}_{i,2})/n$, where $\mathcal{L}_{i,1}$ equals:
\begin{align*}
    -\sum_{\bm{\phi}_i^+} \log \frac{\exp(\bm{\phi}_i \cdot \bm{\phi}_i^+/\tau_{sa})}{\sum_{\bm{\phi}_j} \exp(\bm{\phi}_i \cdot \bm{\phi}_j /\tau_{sa})} & = -\sum_{\bm{\phi}_i^{+}} \bm{\phi}_i \cdot \bm{\phi}_i^+/\tau_{sa} - \sum_{\bm{\phi}_i^+} \log \sum_{\bm{\phi}_j} \exp(\bm{\phi}_i \cdot \bm{\phi}_j /\tau_{sa}).
\end{align*}Here the first term in the above can be rewrite as $\sum_{\bm{\phi}_i^{+}} \|\bm{\phi}_i - \bm{\phi}_i^+\|^2/(2\tau_{sa})-1$ given the normalization $\|\bm{\phi}_p\|=1$ for all pixels $p$. 
Then by the definition $f(\omega_p; \mathbf{x}) = \bm{\phi}_p$ and \eqref{eq: positive alignment}, it is clear that $\mathcal{L}_{i,1}$ induces small $A$ (i.e. good alignment). 

Similar analysis shows $\mathcal{L}_{i,2}$ encourages $\bm{\phi}_i$ to be close to the pre-computed optimal class center $\bm{\nu}_i$ (small $\|\bm{\phi}_i - \bm{\nu}_i\|$). 
The class centers computed from solving \eqref{loss:unif} induces large distance $\|\bm{\nu}_i - \bm{\nu}_j\|$ between centers.
Furthermore, since \eqref{loss:unif} does not involve any data yet, it is immune to long-tailness and can guarantee well-separeted centers for the representation of tail classes. 
Together it holds that $\mathcal{L}_{i,2}$ encourages large $\|\overline{\bm{\phi}_c} - \overline{\bm{\phi}}_{c'}\|$ for $c\neq c'$, or equivalently small $\overline{\bm{\phi}}_c \cdot \overline{\bm{\phi}}_{c'}$, which is exactly the class divergence.

\end{document}